\documentclass[times,twocolumn,final,authoryear]{article}

\usepackage{framed,multirow,natbib}
\usepackage{comment}

\usepackage{textcomp}
\usepackage{graphicx,epsfig,amssymb}
\usepackage{enumerate}
\usepackage{multirow}
\usepackage[caption=false,font=footnotesize]{subfig}

\DeclareGraphicsExtensions{.pdf,.jpeg,.png}

\def\Nys{Nystr{\"o}m }
\def\Nysn{Nystr{\"o}m}

\title{Scalable out-of-sample extension of graph embeddings using deep
  neural networks}
  \author{Aren Jansen, Greg Sell, Vince Lyzinski\\
 \small{Human Language Technology Center of
  Excellence}}

\begin{document}

\maketitle

\begin{abstract}
Several popular graph embedding techniques for representation learning
and dimensionality reduction rely on performing computationally
expensive eigendecompositions to derive a nonlinear transformation of
the input data space.  The resulting eigenvectors encode the embedding
coordinates for the training samples only, and so the embedding
of novel data samples requires further costly computation.  In this paper, we present
a method for the out-of-sample extension of graph embeddings using deep
neural networks (DNN) to parametrically approximate these nonlinear
maps.  Compared with traditional nonparametric out-of-sample extension
methods, we demonstrate that the DNNs can generalize with equal or
better fidelity and require orders of magnitude less computation at test
time.  Moreover, we find that unsupervised pretraining of the DNNs
improves optimization for larger network sizes, thus removing
sensitivity to model selection.
\end{abstract}



\section{Introduction}

Manifold learning is a popular data analysis framework
that attempts to recover compact low-dimensional embeddings of
high-dimensional datasets.  Several manifold learning
algorithms---including ISOMAP~\citep{bib:isomap}, locally linear
embedding~\citep{bib:lle,bib:lle2}, diffusion maps~\citep{Coifman.Lafon:2006A}, and Laplacian
eigenmaps~\citep{bib:BelNiy03}---derive coordinate representations that
encode the local neighborhood structure of an unlabeled data sample.
These techniques have found considerable success in a wide array of
application domains, including computer
vision~\citep{elgammal2004inferring,murphy2009head,he2005face}, speech
processing~\citep{bib:JanNiy06,bib:isa,bib:tsp,tomar2013efficient,sahraeian2013study,Mousazadeh.Cohen:2013A},
and natural language processing~\citep{shi2007text,solka2008text}.
In~\citep{bib:graphembed}, it was shown that these algorithms are all members
of a more general graph embedding framework, in which
the transformations are derived via a generalized eigendecomposition of
the graph Laplacian matrix operator for algorithm-specific graph
construction methodologies.

In their basic form, these graph embedding techniques only provide
transformations of the training samples used to construct the graph.
Thus, even if a large training set is used, computing the output of the
estimated map for a novel test sample is not possible.  To address this
shortcoming, a nonparametric out-of-sample extension technique based on
\Nys sampling was developed that leverages the input and target
representation pairs for each training sample to approximate what the
map would have generated for an arbitrary test
point~\citep{bib:bengio,kumar2012sampling}.  While generally effective,
the \Nys extension is a kernel-based method with time complexity that
scales linearly with the number of training samples.  This increase in 
computational cost is especially problematic because manifold methods
are most effective when provided the benefit of large training sets for
representation learning.  It would be highly beneficial to remove this
trade-off between representation quality and extension feasibility with
a more efficiently scaleable method for out-of-sample
extension.

Neural networks have long been known to be a powerful learning framework
for classification and regression, capable of distilling large training
sets into efficiently evaluated parametric models, and thus are a natural
choice for modeling manifold embeddings.  In their seminal
paper, Hornik \emph{et al.}~\citep{hornik1989multilayer} proved that
feedforward neural networks can approximate a virtually arbitrary
deterministic map between high-dimensional spaces, indicating that they
would also be ideally suited for our out-of-sample extension problem.
However, there are two caveats for the use of neural networks as 
universal approximators: (i) there must be sufficient hidden units
(i.e. sufficient model parameters), which in turn require additional data samples
for training without overfitting; and (ii) the non-convexity of the
objective function grows with the number of model parameters, making the
search for reliable global solutions increasingly difficult.

With these considerations in mind, we explore the application of recent
advances in deep neural network (DNN) training methodology to the
out-of-sample extension problem.  First, by stabilizing the Lanczos
eigendecomposition algorithm, we are able to produce exact graph
embeddings for training sets with millions of data samples.  This
permits an extensive study with deeper architectures than have been
previously considered for the task.  Second, motivated by success in the
supervised classification setting~\citep{bengio2009learning,bengio2007greedy}, we consider unsupervised DNN
pretraining procedures to improve optimization as our larger training
samples support commensurate increases in model complexity.  

In the work that follows, we compare
the performance of our parametric DNN approach against a \Nys sampling
baseline, both in terms of approximation fidelity and test runtime.  We
find DNNs to match or outperform the approximation fidelity of the \Nys method for
all training sample sizes.  Furthermore, since the DNN approach is
parametric, its test-time complexity for fixed network size is constant
in the training sample size, producing orders-of-magnitude speedup over
\Nys sampling for larger training sizes.  The remainder of this paper is organized as follows. We begin with an overview of prior work in out-of-sample
extension for graph embeddings.  We then describe the strategy for stabilizing eigendecompositions
for large training sets, followed by a description of the process for training our DNN out-of-sample extension
to approximate the embedding for unseen data.  Finally, we analyze the reconstruction accuracy and 
computation speed of both the \Nys baseline and the DNN approach. 

\section{Prior Work}
\label{sec:prior}

The most popular methods for extending graph embeddings to unseen data
have been based on \Nys sampling~\citep{bib:bengio,kumar2012sampling}, and
thus they will serve as the baseline in our experiments.  This is a
nonparametric, kernel-based technique that approximates the embedding of
each test sample by computing a weighted interpolation of the embeddings
for training samples that were nearby in the original input space.
Formally (see~\citep{bib:bengio} for details), let $X =
\{x_1,\dots,x_n\}$ be the set of graph embedding training samples, where
each $x_i \in \mathbb{R}^d$.  Let $\mathcal{L}$ be the symmetric, normalized graph
Laplacian operator defined for the set $X$ such that $\mathcal{L} =
I-D^{-1/2}AD^{-1/2}$, where $A_{i,j}=K(x_i,x_j)$ for some positive
semidefinite kernel function $K$ that must be specialized for each specific
graph embedding algorithm; and $D$ is the diagonal matrix defined 
via $D_{i,i}=\sum_j A_{i,j}$.  Let the spectral decomposition of the normalized Laplacian be 
denoted as $\mathcal{L}=U\Sigma U^{T}$, where the diagonal entries of $\Sigma$ are non-increasing.
The $d'$-dimensional embedding of $X$ is then provided by the first $d'$ columns of $U$, which we
shall denote as $U_{d'}$.  Stated simply, the embedding of $x_i$ is given by the $i$-th row of $U_{d'}$.

To embed an out-of-sample data point $x \in \mathbb{R}^d$ via the \Nys extension, the $p$-th 
dimension of the extension $y_p(x)$ is given by
\begin{equation}
\label{eq:nys}
  y_p(x) = \frac{1}{\lambda_p} \sum_{i=1}^n v_{pi} K(x,x_i).
\end{equation}
\noindent
We see that the complexity of this extension is linear in the size of
the training set. In practice, approximate nearest neighbor techniques
can be used to speed this up with minimal loss in fidelity (the
implementation we benchmark uses k-d tree for this purpose), but the algorithmic
complexity still increases with training size.  Finally,
note that a nearly equivalent formulation based on reproducing Kernel
Hilbert space theory was presented in~\citep{bib:BelNiySin06}, where the
kernelization was introduced into the objective function before the
eigendecomposition is performed.  This formulation has the same
scalability limitations as the \Nys extension.  These computational 
difficulties motivate our exploration of DNNs to model embeddings for 
out-of-sample extension.

Traditional neural networks have also been considered for out-of-sample
extension in the past in two limited studies involving small datasets
and model architectures~\citep{gong2006neural,chin2007extrapolating}.
The idea was introduced in~\citep{gong2006neural}, but the study failed
to include a meaningful quantitative evaluation.  The experiments
in~\citep{chin2007extrapolating}, which predated the advent of recent
deep learning training methodologies, found neural networks to be one of
the worst performing methods.  However, with a similar motive of
computational efficiency, \citep{gregor2010learning} explored the use of
DNNs for approximating expensive sparse coding transformations and
produced more compelling results.  

\section{A Scalable Out-of-Sample Extension}

A truly scalable out-of-sample extension must simultaneously consume a
large amount of training data for detailed modeling and provide a
test-time complexity that does not strongly depend on that training set
size.  The nonparametric nature of the \Nys method leads to a linear
dependence on the training set size (logarithmic if kernel
approximations are implemented) and thus can get bogged down as we feed
more data to the graph embedding training.  We begin this section with a
simple trick for the eigendecomposition of large graph Laplacians, which
permits larger training sets and motivates the need for more
computationally efficient extension methods.  This is followed by a
presentation of the deep neural network architecture we propose to
efficiently extend the embedding to arbitrary test points.

\subsection{Stabilizing the Eigendecomposition}
\label{sec:bigeigs}

In~\citep{golub2012matrix}, it is suggested that the stability of the
Lanczos eigendecomposition algorithm can be greatly increased (and memory
requirements consequently reduced) by reformulating the eigenproblem to
recover the largest eigenvalues.  We can exploit this by observing that
if $v$ is an eigenvector of $\mathcal{L}$ with eigenvalue $\lambda$, then $v$ is
also an eigenvector of $\widetilde{\mathcal{L}} = I-\mathcal{L}$ with eigenvalue $1-\lambda$
(which is guaranteed to be less than or equal to 1).  Thus, with this
small redefinition of the eigenproblem, we can recover the same
eigenvectors by considering the largest eigenvalue criterion.  Note that
when using the ARPACK implementation, a similar effect can also be
accomplished by searching for the smallest \emph{algebraic} eigenvalues
of $\mathcal{L}$ directly.

While this trick is by no means a fundamental theoretical innovation on
our part, its effects have proven dramatic.  Our past efforts to solve
for the smallest magnitude eigenvalues of the graph Laplacian exceeded
our hardware memory limits when our graphs reached the order of 100,000
nodes and 1 million edges.  Employing this simple trick, we have now
succeeded in processing graphs with order 100 million nodes and order 10
billion edges on conventional hardware, stably solving for the top 100
eigenvectors in a few days using 32 cores and 0.5 TB of RAM.  This
problem size even exceeds what was reported using approximate singular
value decomposition solvers in the past~\citep{talwalkar2008large}.  For
the 1.5 million node graphs we consider in our experiments described
below, this method was more than adequate for our (offline) embedding
training needs.  

\subsection{Deep Neural Network Methodology}

Solving the eigenvalue problem produces a $d'$-dimensional (exact) embedding $z_i \in
\mathbb{R}^{d'}$ for each $x_i \in X$. Rather than viewing new data points as out-of-sample
points whose mapping is estimated with interpolation, we instead seek to estimate the mapping
(from $x$ to $z$) itself.  To this end, we now consider 
feedforward neural architectures with $N$ hidden layers, each containing $M$ hidden units.
The $l$-th hidden layer nonlinearly maps the output of the previous
layer $h_{l-1}$ to a new hidden representation $h_l \in \mathbb{R}^M$
according to 

\begin{equation}
  h_l = \sigma(W_lh_{l-1} + b_l),
\end{equation}

\noindent
where each $W_l$ is a parameter matrix, $b_l$ is a bias column vector, and
$\sigma$ is the activation function (which we set to $\tanh$ in our
experiments).  The input $h_0$ to the first layer is a point $x$ in our
input space $\mathbb{R}^d$, while $W_1 \in \mathbb{R}^{M\times d}$, $W_l
\in \mathbb{R}^{M\times M}$ for $l \in \{2,\dots,N\}$, and $b_l \in
\mathbb{R}^{M}$ for $l \in \{1,\dots,N\}$.  The output $h_l$ of these
$N$ hidden layers are finally transformed into a corresponding point
$y(x) \in \mathbb{R}^{d'}$ according to 

\begin{equation}
\label{eq:fwdpass}
y(x) = W_{N+1}h_N + b_{N+1}, 
\end{equation}

\noindent
where $W_{N+1} \in \mathbb{R}^{d'\times M}$ and $b_{N+1} \in
\mathbb{R}^{d'}$ are the decoding weight matrix and bias column vector,
respectively.  The training objective is to solve for parameters
$\Theta^* = \{W_1^*,\dots,W_{N+1}^*,b_1^*,\dots,b_{N+1}^*\}$ that
minimize mean squared error between the exact and predicted embedding
pairs:

\begin{equation}
\label{eq:dnnopt}
\Theta^* = \arg \min_{\Theta} \frac{1}{n} \sum_{i=1}^n \| z_i - y(x_i) \|^2.
\end{equation}

\noindent
This is generally accomplished with backpropagation and stochastic
gradient descent optimization.

It is critical that the training procedure safeguards against
overfitting to the training sample, especially as deeper architectures
are required in order to approximate the detailed graph embeddings we
wish to extend.  Our training procedure, which was also employed
in~\citep{kamperunsupervised} for a different application, has two steps:
(i) unsupervised stacked autoencoder pretraining that uses $X$ only, and
(ii) supervised fine-tuning using the $(x_i,z_i)$ pairs as training
inputs and targets.  

\subsubsection{Unsupervised Pretraining}

When the input and output targets are the same, our deep network
architecture reduces to a stacked autoencoder (SAE) with $N$ encoding
layers and one decoding layer.  Thus, to initialize model parameters of
the DNN to approximately recover the identity mapping, we consider the
unsupervised SAE pretraining
procedure~\citep{bengio2009learning,bengio2007greedy}.  Here, we
introduce one hidden layer at a time, performing several epochs of
stochastic gradient descent to minimize mean squared error between our
training samples and themselves at each intermediate network depth.  As
we add each new layer, we discard the linear decoding weights from the
previous optimization, use the previous hidden representation as input
to the new hidden layer, and reoptimize all layer parameters.  Early
stopping is used to prevent exact recovery of the identity map for each
layer. 

\subsubsection{Supervised Fine-Tuning}

Using the above layer-wise pretraining procedure, we now have
initialized all parameters in the network.  It only
remains to perform several epochs of stochastic gradient descent to
reoptimize network parameters to minimize mean squared error between the
exact and predicted embedding pairs according to
Equation~(\ref{eq:dnnopt}). After this training is complete, the DNN-based
out-of-sample extension $y(x)$ for an arbitrary point $x \in
\mathbb{R}^d$ can be efficiently computed using the standard neural
network forward pass defined by Equation~(\ref{eq:fwdpass}).  This
amounts to $N+1$ matrix-vector multiplies and vector additions,
plus an evaluation of $\sigma$ for each hidden unit.  For a fixed
network architecture, this computation is constant in the number of
training samples.

\begin{figure*}[t]
\centering
\begin{tabular}{rrrrl}
\subfloat[\Nysn, $n=1,000$]{\includegraphics[width=0.435\columnwidth]{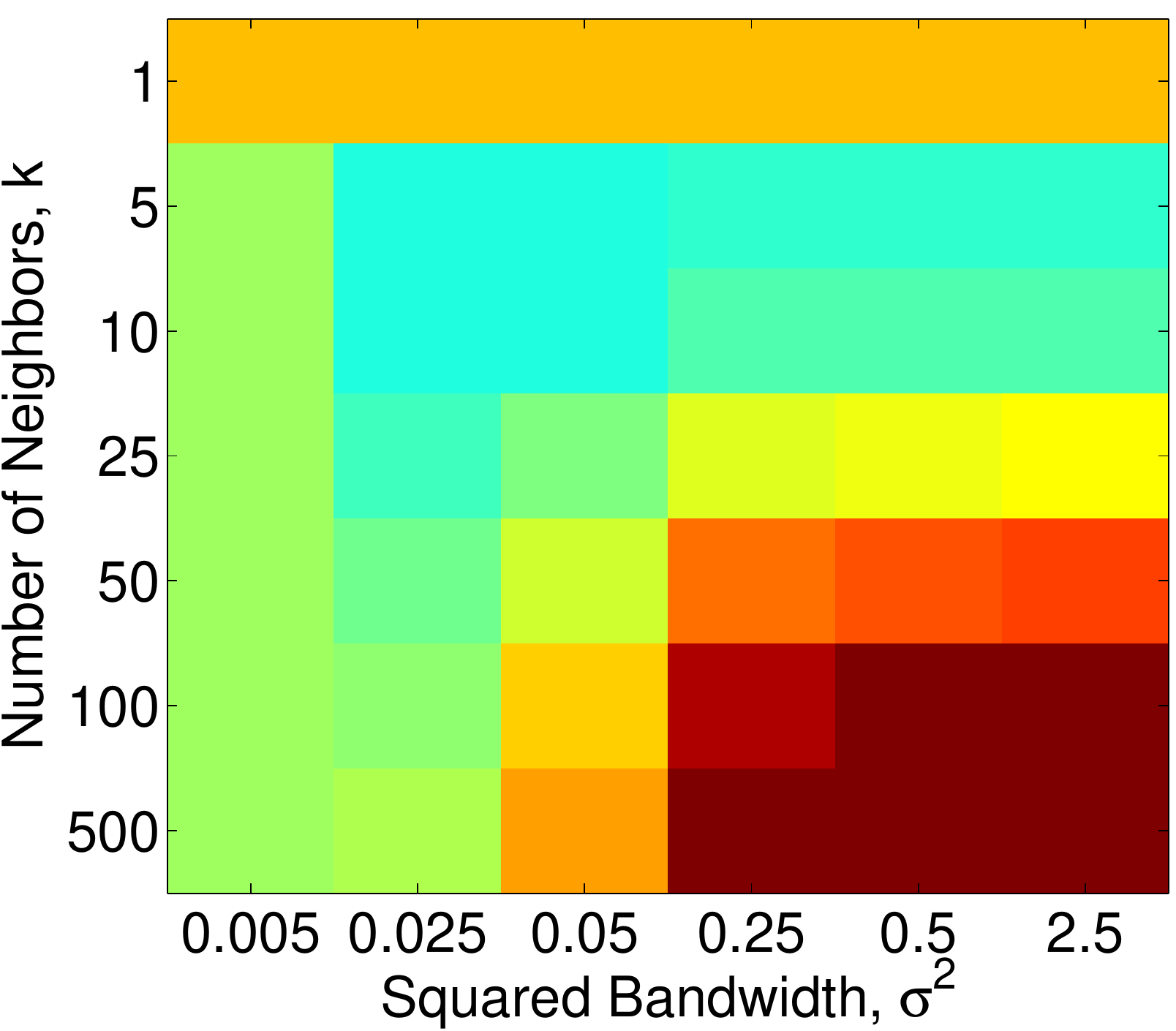}} &
\subfloat[\Nysn, $n=10,000$]{\includegraphics[width=0.435\columnwidth]{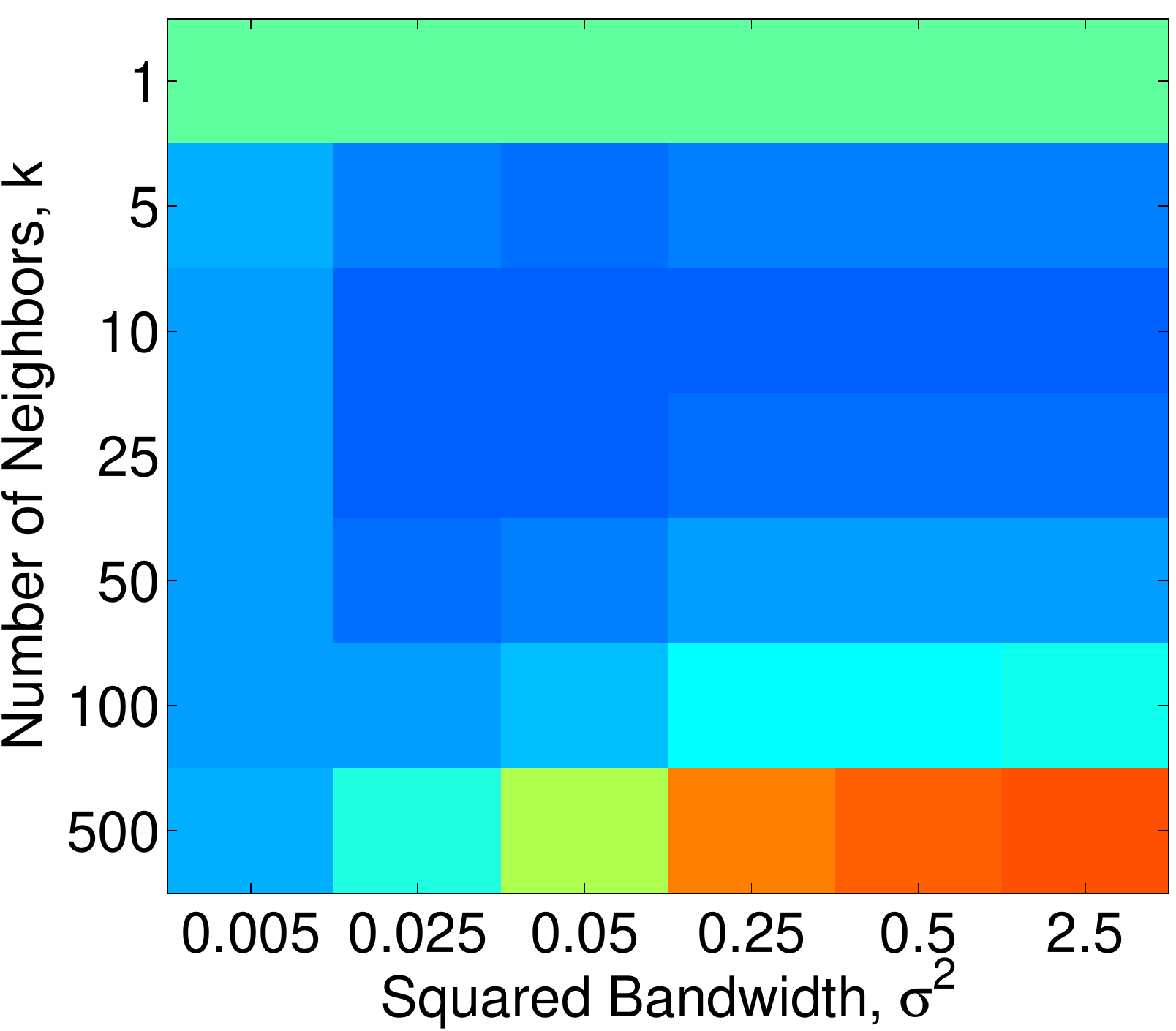}} &
\subfloat[\Nysn, $n=100,000$]{\includegraphics[width=0.435\columnwidth]{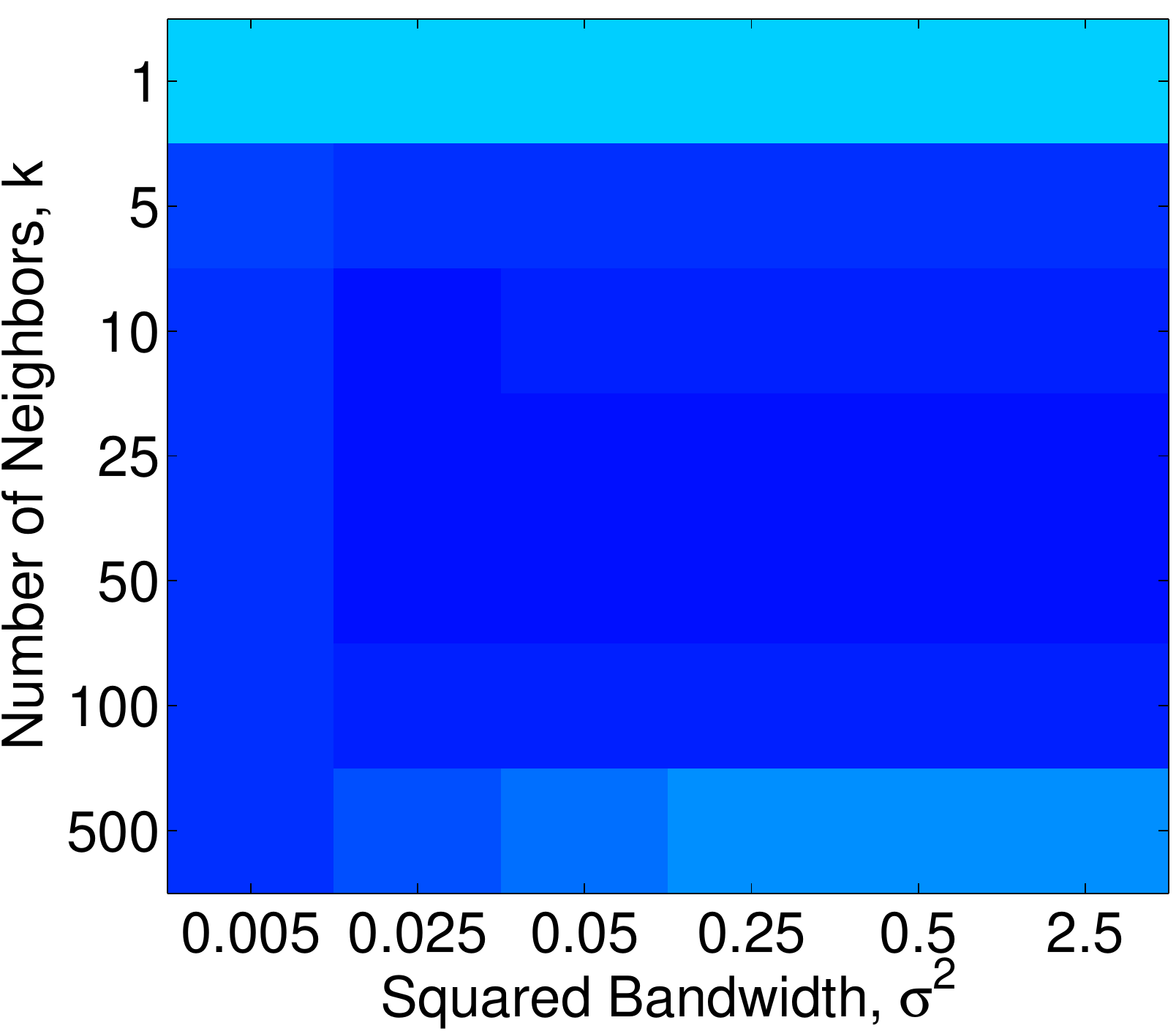}} &
\subfloat[\Nysn, $n=1,100,000$]{\includegraphics[width=0.435\columnwidth]{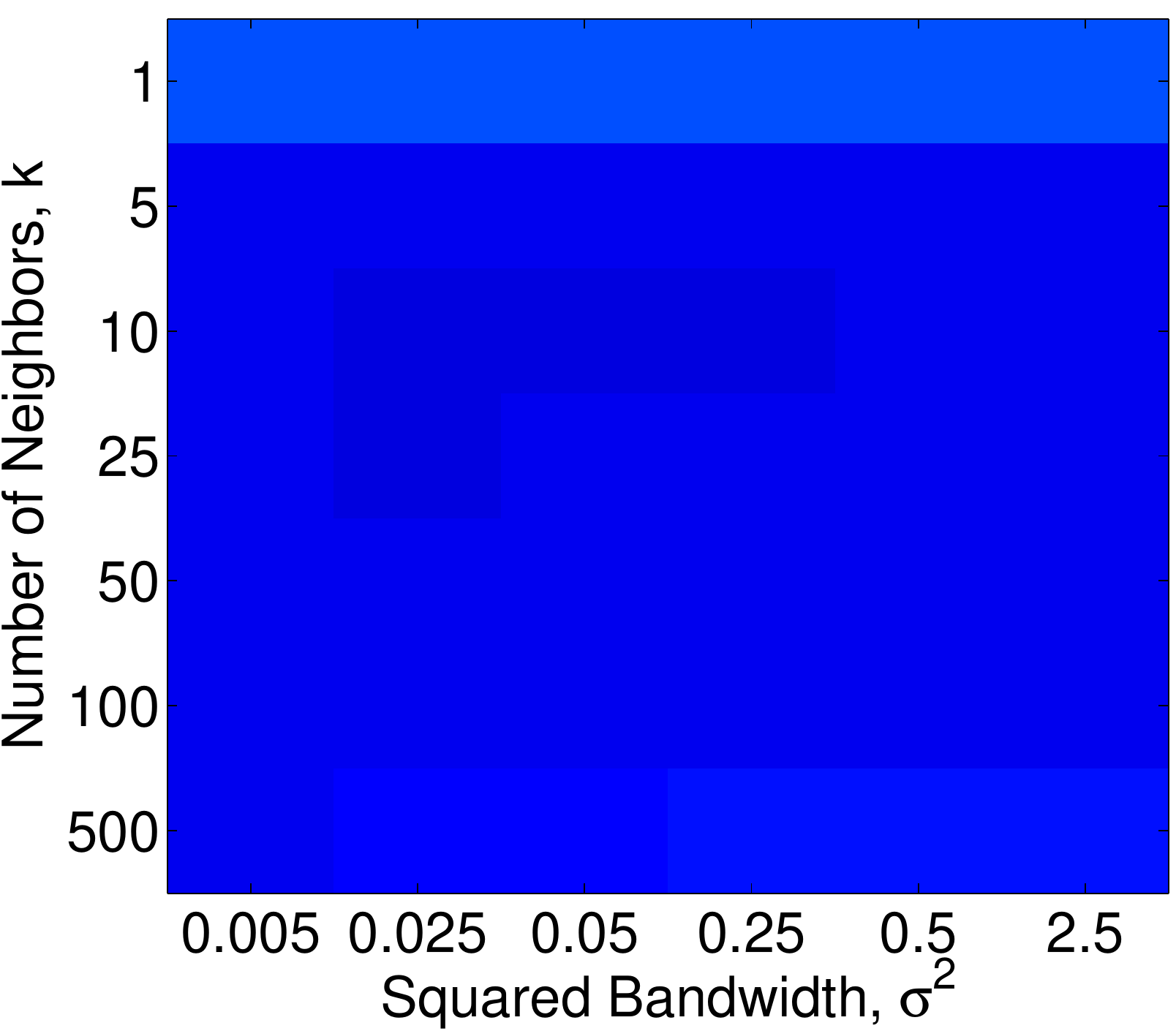}} &
\vspace{0.5cm}
\captionsetup[subfigure]{labelformat=empty}
\multirow{2}{*}[2cm]{\subfloat[NRMSE]{\includegraphics[width=0.075\columnwidth]{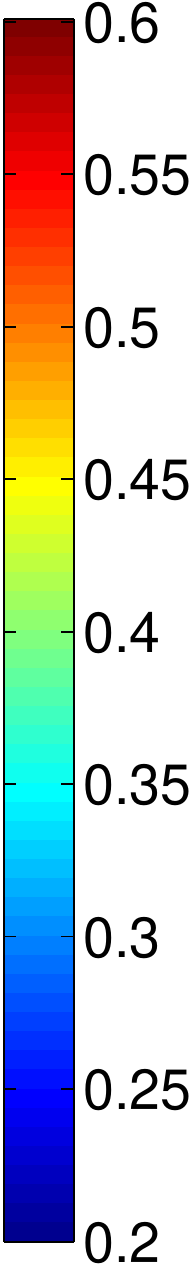}}}
\\ 
\setcounter{subfigure}{4}
\subfloat[DNN, $n=1,000$]{\includegraphics[width=0.413\columnwidth]{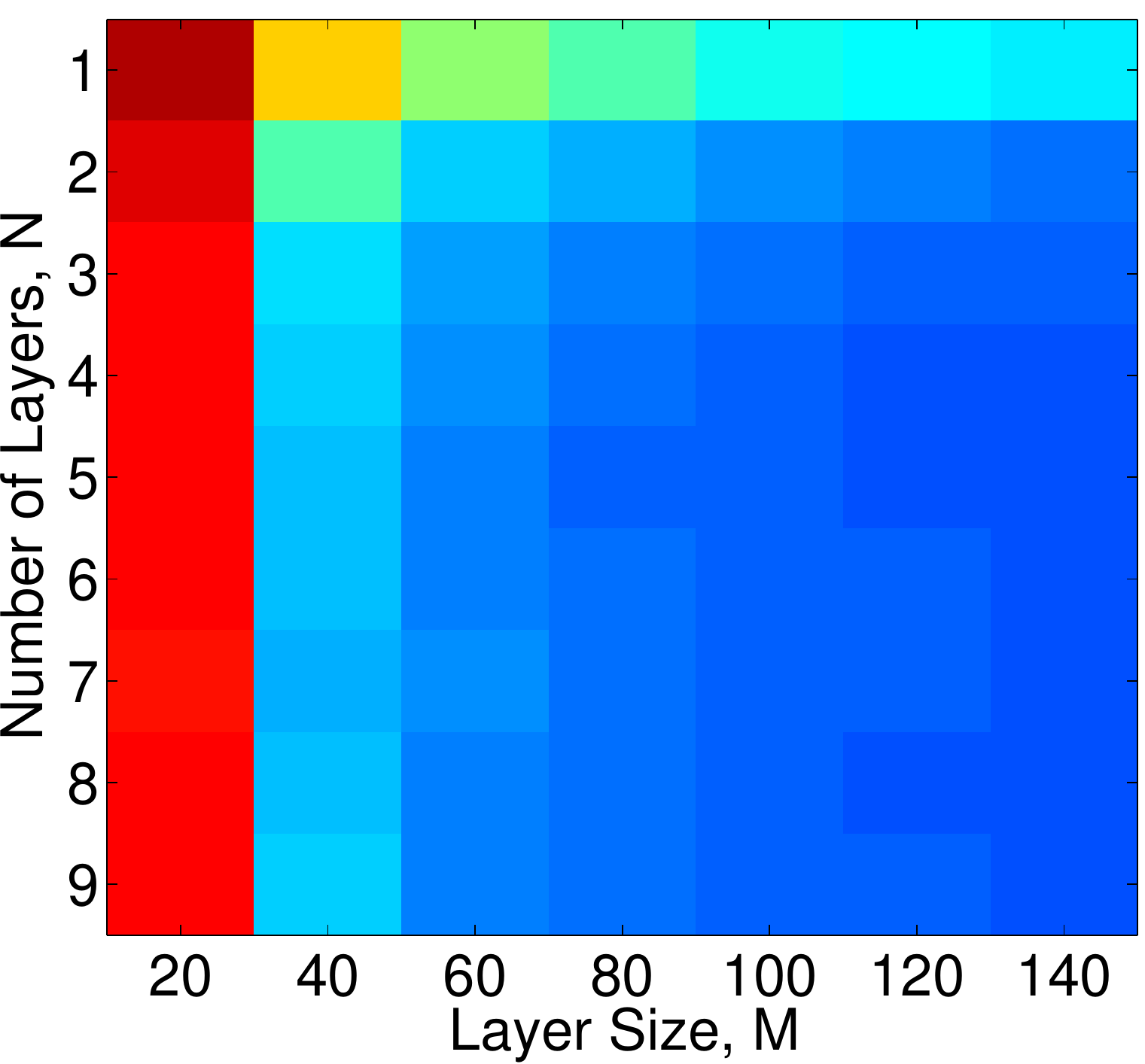}} &
\subfloat[DNN, $n=10,000$]{\includegraphics[width=0.413\columnwidth]{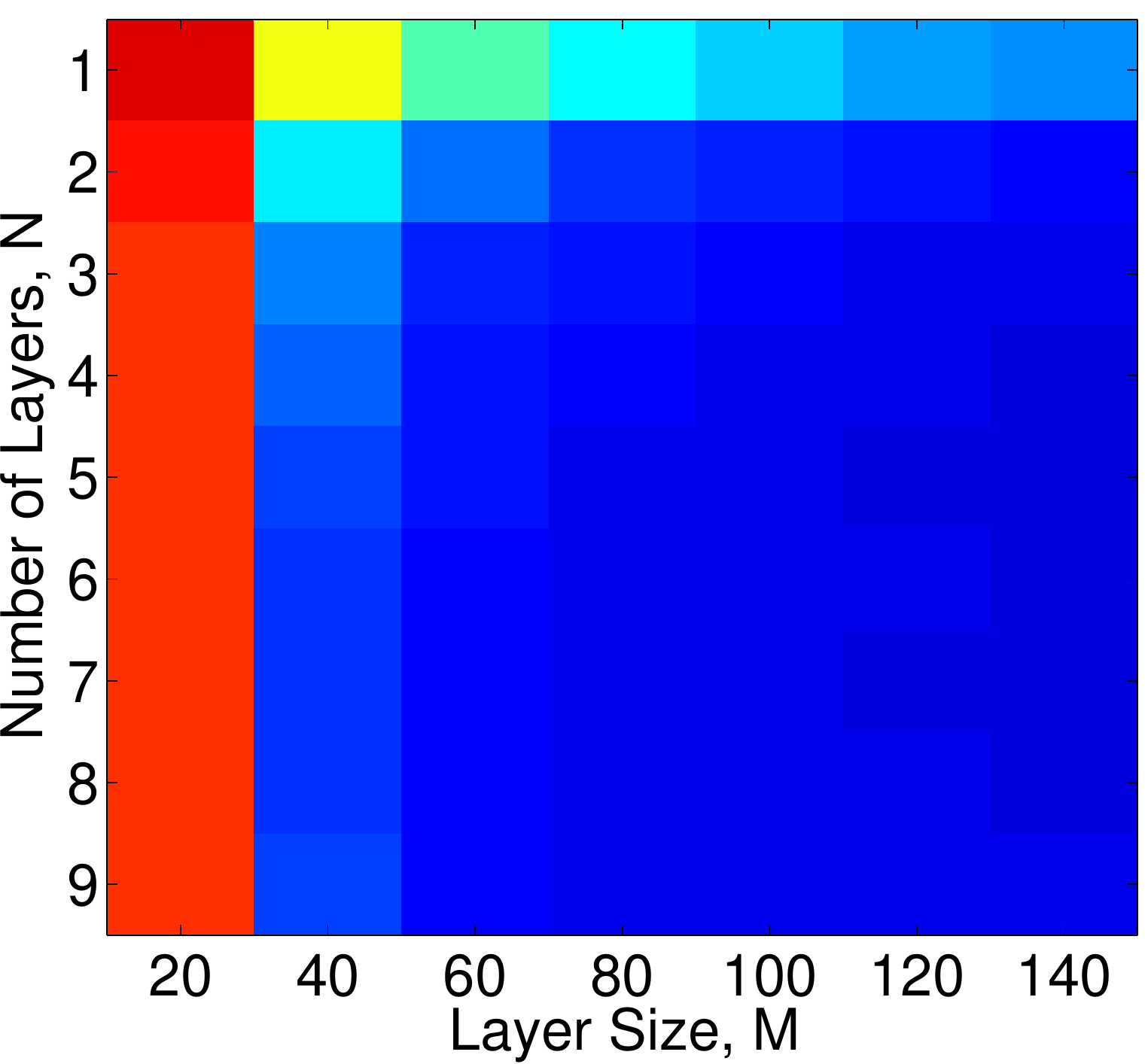}} &
\subfloat[DNN, $n=100,000$]{\includegraphics[width=0.413\columnwidth]{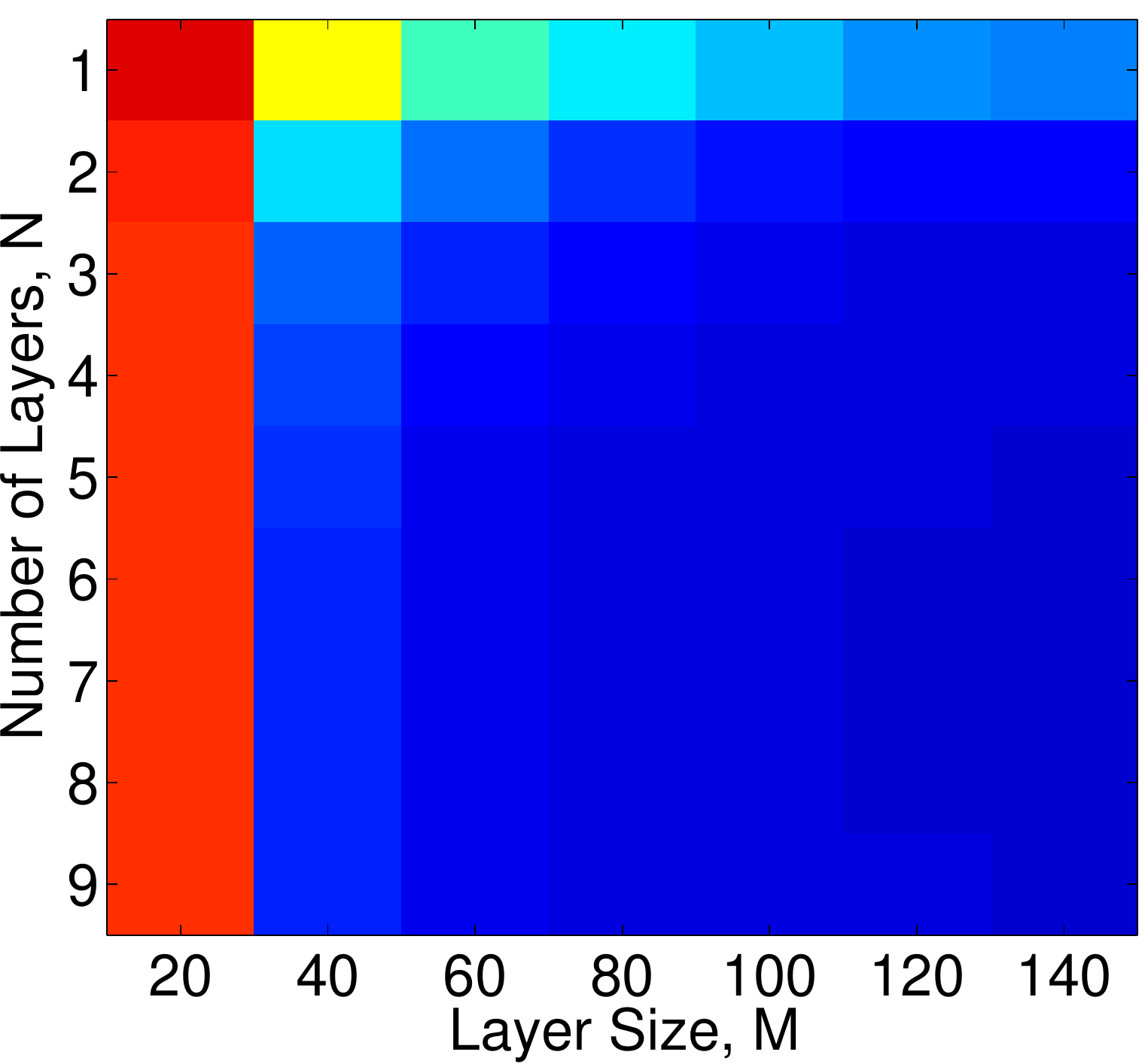}} &
\subfloat[DNN, $n=1,100,000$]{\includegraphics[width=0.413\columnwidth]{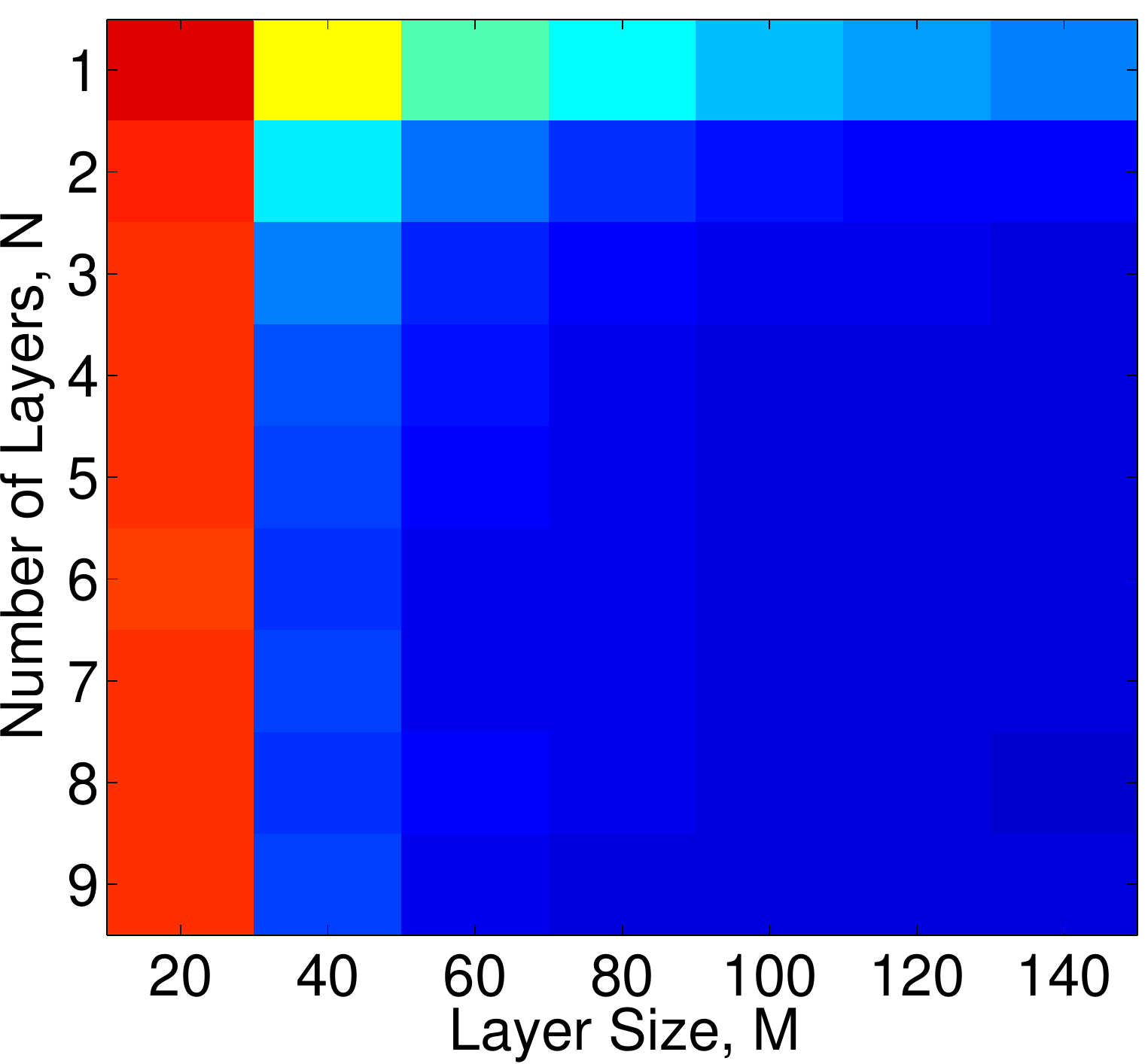}}&
\end{tabular}
\caption{Heat maps showing the NRMSE values for all \Nys (a-d) and DNN (e-h)
  hyperparameter settings, for each of the four training set sizes. The
  same color scale is used for all images. }
\label{fig:params}
\end{figure*}

\section{Experiments}

We choose speech as the application domain for our scalability study for
two reasons:  (i) a single
hour of audio recordings typically produce 360,000 high-dimensional data
samples known as frames, and so large datasets are readily available; and (ii) manifold learning techniques have been
shown to learn representations effective for keyword discovery and
search~\citep{bib:isa,bib:tsp,jansen2013summary}.  We use the TIMIT corpus for evaluation, 
given the past success of manifold embeddings for speech recognition on that data~\citep{bib:isa}.  It consists of over 4 hours
of prompted American English speech recordings, and is split into a training set consisting of roughly 1.1 million data
samples ($\sim$3 hours) and a test set of roughly 400,000 data
samples ($\sim$1 hour).  There is no speaker overlap between the
two sets.

\subsection{Evaluation Methodology}

Our goal is to measure the approximation fidelity of the out-of-sample
extension of the test set against a reference embedding that is
independent of the extension method.  Thus, our strategy is to perform
an exact graph embedding of the entire corpus (train+test) using the
method described in Section~\ref{sec:bigeigs}; we call these the
\emph{reference embeddings}.  We define each out-of-sample extension
using input frames and corresponding reference embeddings from the
training set.  This allows us to use reference embeddings for the
training set to approximate embeddings for the test set for comparison
against the true reference embeddings of the test set.   We measure
approximation fidelity in terms of normalized root mean squared error
(NRMSE) between the predicted and reference test embeddings; here, the
normalization constant is taken to be the root mean squared error
between the test set reference embeddings and a random permutation of
those same samples.  Thus, a perfect out-of-sample extension will have
NRMSE of 0, while a extension that is a random mapping with the same
empirical output distribution will have NRMSE of 1. In addition to
defining the extensions with the entire training set (1.1 M samples), we
also consider the utility of random subsets of sizes 1,000, 10,000, and
100,000.  However, we use the same reference embeddings for all
experiments.

Our input features are 40-dimensional, homomorphically smoothed, mel-scale, log
power spectrograms (40 equally spaced mel bands from 0-8 kHz), sampled
every 10 ms using a 25 ms Hamming window.  We construct the graph Laplacian using a symmetrized
10-nearest neighbor adjacency graph with cosine distance as the metric
and binary edge weights.  This amounts to a Laplacian eigenmaps
specialization of the graph embedding framework.  Finally, we keep the
40 eigenvectors with the largest eigenvalues to produce a graph
embedding with the same dimension as the input space.  While our present
focus is on out-of-sample extension fidelity, it is relevant to note
that the reference embeddings match the best downstream performance
reported in~\citep{bib:tsp}, which used an identical embedding strategy.

For the baseline \Nys method, we compute Equation~(\ref{eq:nys}) using a
nearest neighbor approximation to a radial basis function (RBF) kernel.
This approximation is facilitated by preprocessing the training samples
into a k-d tree data structure (as implemented in \texttt{scikit-learn})
for efficient retrieval of near-neighbor sets.  Note that we tried
matching the \Nys kernel to that used in the graph construction (i.e.,
using binary weights) as prescribed by~\citep{bib:bengio}, but it
performed substantially worse than introducing RBF weights.  We consider
kernel squared-bandwidths $\sigma^2 \in \{0.005, 0.025, 0.05, 0.25, 0.5,
2.5\}$ and number of neighbors $k \in \{1,5,10,25,50,100,500\}$. 

For our DNN method, we consider $L \in \{1,2,\dots,9\}$ hidden layers,
and for each depth we evaluate layer sizes of $M \in \{20,40,60,
80,100,120,140\}$ hidden units.  For pretraining, we use the entire
training set and optimize each layer for 15 epochs of stochastic
gradient descent.  Following the prescription
in~\citep{kamperunsupervised}, we use a learning rate of
$2.5\times10^{-4}$ and minibatches of 256 samples.  For supervised
fine-tuning, we increase the number of epochs for the smaller training
sets such that the total number of examples processed is roughly fixed
(5 epochs for the full train set of 1.1 million samples, 50 epochs for
100,000 samples, etc) to ensure adequate convergence.  Also
following~\citep{kamperunsupervised}, for fine-tuning we use a learning
rate of $4\times 10^{-3}$, but found a smaller minibatch of size 50
improved convergence.  We use the \texttt{Pylearn2}
toolkit~\citep{goodfellow2013pylearn2} for all DNN experiments.

\begin{table*}
  \caption{Test set NRMSE and runtime (in seconds, averaged over several
    trials) for \Nys with optimal
    hyperparameters, a small DNN ($N=2,M=60$), and a large DNN ($N=5,M=140$).}
\label{tab:best}

  \begin{center}
  \begin{tabular} {  c | c | c | c | c | c | c  }
    \hline
           Train      & \multicolumn{2}{c|}{\Nys (optimal)} &
    \multicolumn{2}{c|}{Small DNN} & \multicolumn{2}{c}{Large DNN} \\ \cline{2-7}
    Size   & NRMSE & Time   & NRMSE    & Time & NRMSE & Time
    \\ \hline 
    1k     & 0.36  & 25     & 0.33     & 5.4  & 0.28  & 33 \\
    10k    & 0.29  & 240    & 0.29     & 5.4  & 0.24  & 33 \\
    100k   & 0.25  & 2,200  & 0.29     & 5.4  & 0.23  & 33 \\
    1.1M   & 0.24  & 12,000 & 0.29     & 5.4  & 0.23  & 33 \\ \hline
  \end{tabular}
  \end{center}
  \vspace{-0.5cm}
\end{table*}

\subsection{Hyperparameter Sensitivity}

First, we consider the NRMSE performance as we vary the amount of
training samples used by the out-of-sample extension.  We drew random
subsets of the 1.1 million sample training set of sizes 1000, 10,000,
and 100,000.  Each of these subsets was used for computation of
Equation~(\ref{eq:nys}) in the \Nys method and for supervised
fine-tuning in the case of the DNN method.  Table~\ref{tab:best} lists
for each training subset size the NRMSE and test runtime for (i) \Nys
using the optimal set of hyperparameters, (ii) a smaller DNN with 2
layers of 60 units each, (iii) a larger DNN with 5 layers of 140 units.
We see that the larger DNN matches or outperforms the \Nys method for
all training set sizes, demonstrating the power of deep learning for
accurately approximating complex nonlinear functions.  While the small
DNN matches \Nys for the 2 smaller training sets, it does not have
sufficient parameters to keep pace as more training data becomes
available.

These results emphasize the importance of each method's sensitivity to
the choice of hyparameters that specify the complexity of the
out-of-sample extension function.  This is especially true for fully
unsupervised representation learning settings, where cross-validation
may not be possible.  To explore this consideration, for the \Nys
method, we vary the number of (approximate) nearest neighbors that
contribute to each test sample, as well as the kernel bandwidth.  For
the proposed DNN method we vary both the number of layers ($N$) and the
number of hidden units per layer ($M$).  

Figure~\ref{fig:params} shows
heatmaps indicating the NRMSE for all hyperparameters considered for the
two methods for the four training set sizes.  For the \Nys method,
performance is relatively stable across the range of kernel bandwidths,
but it is more sensitive to the number of neighbors used in the
computation.  Moreover, the optimal number of neighbors increases as
more training data is available, necessitating some amount of parameter
tuning to achieve optimal approximation.  

The DNN approach reaches
near-optimal performance for all training set sizes provided we include
at least 4 layers with at least twice as many units than the input
dimension.  Critically, there is no performance penalty for overshooting
the network size (other than increased forwardpass runtime, as we
discuss below).  This suggests the DNN extension would require less
tuning than \Nys method to achieve optimal approximation in new
applications.

\subsection{The Effect of Pretraining}

\begin{figure}[t!]
\centering
\subfloat[1,000 Training Pairs]{\includegraphics[width=.8\columnwidth]{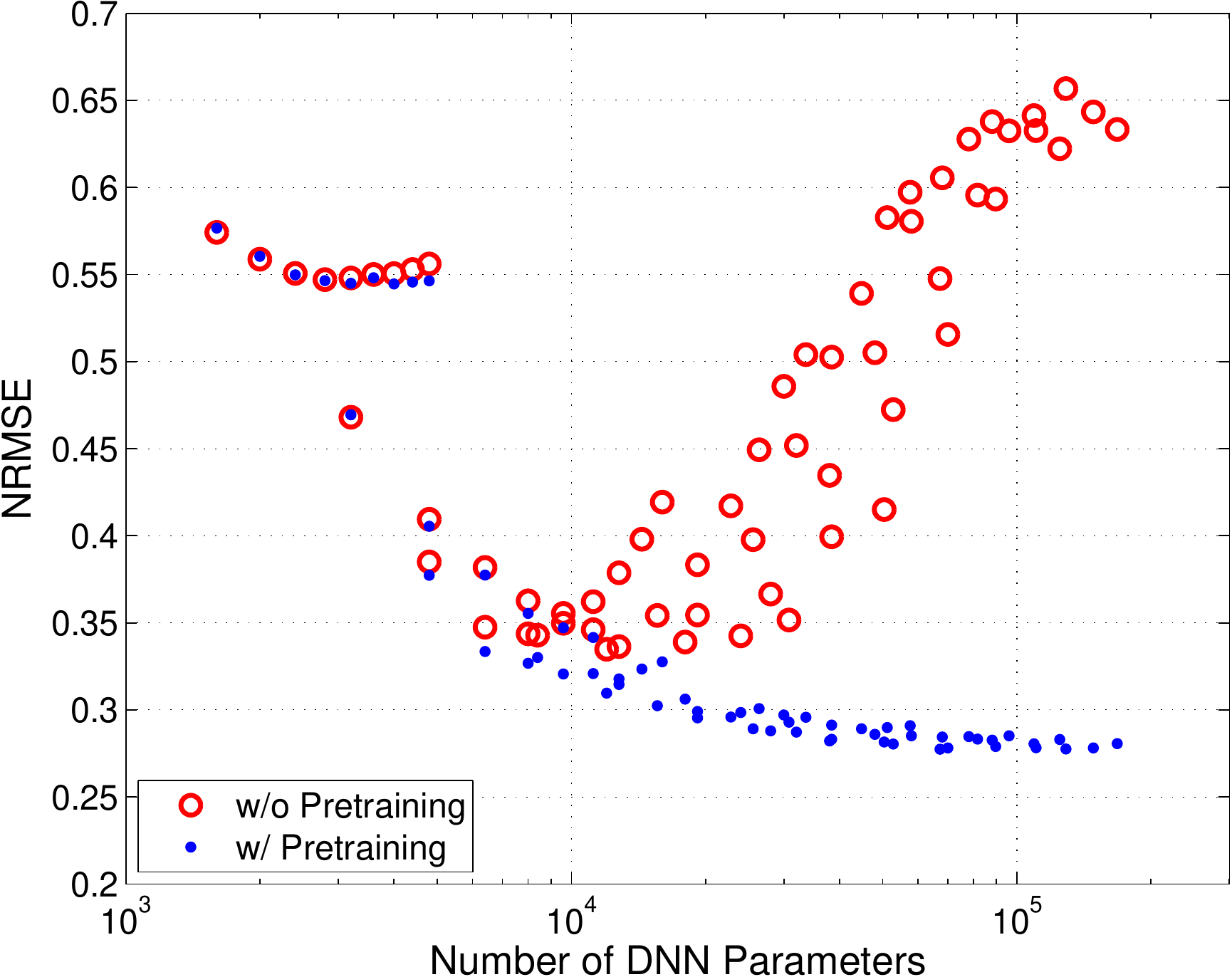}} \\
\subfloat[1,100,000 Training Pairs]{\includegraphics[width=.8\columnwidth]{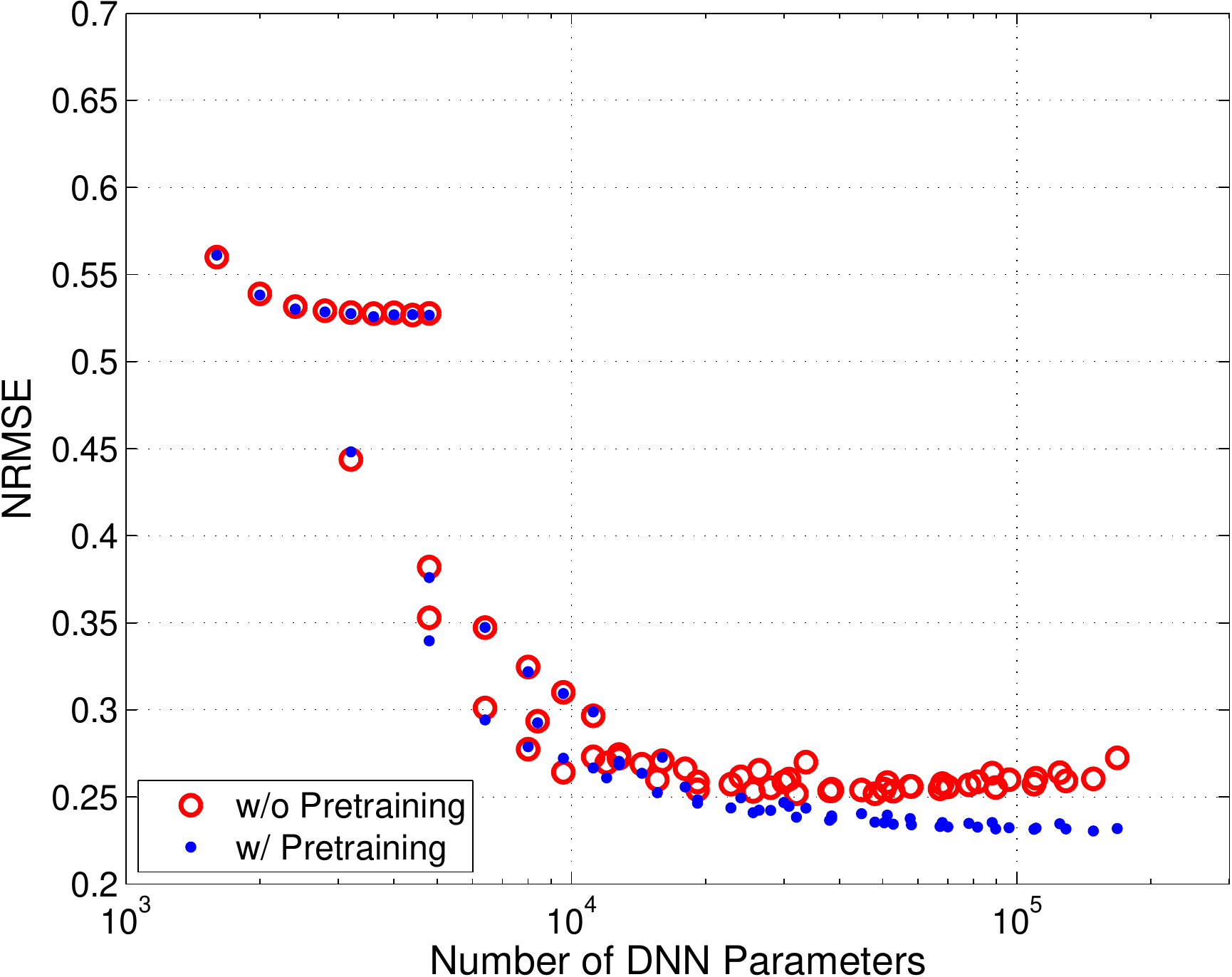}}
\caption{Approximation fidelity vs. number of DNN parameters for (a) 1,000
  and (b) 1.1 million training set sizes, both with and without unsupervised
  pretraining.}
\label{fig:pretrain}
\end{figure}

In typical machine learning scenarios, increasing the number of model
parameters for a fixed training set size opens the method up to
overfitting and poor generalization.  The trends for the DNN methods in
Figure~\ref{fig:params} defy this conventional wisdom, with no loss in
approximation fidelity for any training corpus size as we move to deeper
and wider network architectures.  Indeed, the unsupervised pretraining
is responsible for regularizing the parameter estimates, preventing
overspecialization to even the smallest training set considered.  This
can be seen most clearly in the scatterplots in
Figure~\ref{fig:pretrain}, where each dot represents a single model
architecture.  As the total number of parameters increases, the test
NRMSE of the pretrained models decays roughly monotonically.  The same
architectures without pretraining track similarly for smaller model
sizes, but, due to overfitting, the test performance degrades as more parameters are
made available.  This behavior is especially clear
in the case of limited training examples, though we see the beginnings of the same
effect for the largest architectures even in the presence of the full training set.  

\subsection{Test Runtimes}

Finally, we consider the computational efficiency of applying the two
out-of-sample extension methods to a test corpus.   Table~\ref{tab:best}
lists the NRMSE values and corresponding test times in seconds (i.e.,
the time taken to extend the embedding to the entire 400,000 sample test
set) for the \Nys method with optimal hyperparameters and two DNN
architectures.  As expected, the runtimes for the nonparametric \Nys
method increase as the training sample gets bigger, since nearest
neighbor retrieval remains expensive even when using the k-d tree data
structure.  Meanwhile, the DNN runtimes are virtually constant for a
fixed number of parameters.  The small DNN can consume the full training
set and produce extensions over 4 times faster than the
\Nys method with the smallest training set, while simultaneously reducing 
NRMSE by 20\% relative.  Moreover,
 the best DNN roughly matches the approximation fidelity of the
best \Nys NRMSE, and the DNN accomplishes it $\sim$350 times faster.
For all training sizes tested, DNN extensions can provide signficant
speedup without any sacrifices in fidelity, and, in many cases, improve both speed
and fidelity.

For speech processing applications, where interactivity is often
critical, even our largest networks can process test samples faster than
real-time.  Moreover, with the large DNN consisting of 5 layers of 140
hidden units, optimal NRMSE is achieved at speeds 120 times faster than
real-time using a single processor. This is comparable to the extraction
speed of traditional acoustic front-ends in state-of-the-art
implementations~\citep{povey2011kaldi}.

\section{Conclusion}

In this work, we used modern deep learning methodologies
to perform out-of-sample extensions of graph embeddings.  
Compared with the standard \Nys sampling-based out-of-sample
extension, the DNNs approximate the embeddings with higher fidelity and
are substantially more computationally efficient.  Using unsupervised
pretraining for parameter initialization improves DNN generalization,
making our DNN approach highly stable across a wide variety of
hyperparameter settings.  These results support deep neural networks
with unsupervised pretraining as an ideal choice for out-of-sample
extensions of learned manifold representations.

\section*{Acknowledgment}
The authors would like to thank Herman Kamper of the University of
Edinburgh for providing his correspondence autoencoder tools, on which we
based our neural network implementation.

\end{document}